\documentclass[conference]{IEEEtran}

\usepackage[utf8]{inputenc} 
\usepackage{cite}
\usepackage{amsmath,amssymb,amsfonts}
\usepackage{graphicx}
\usepackage{url}

\usepackage{phonetic}
\usepackage{tipa}
\usepackage{tipx} 

\usepackage{multirow}
\usepackage{booktabs}
\usepackage{tabularx}
\usepackage{siunitx} 

\usepackage{xcolor} 

\usepackage{newunicodechar}
\newunicodechar{ɑ}{\textscripta}
\newunicodechar{æ}{\ae}
\newunicodechar{ɐ}{\textturna}
\newunicodechar{ɒ}{\textinvscripta}
\newunicodechar{ə}{\textschwa}
\newunicodechar{ɛ}{\textepsilon}
\newunicodechar{ɝ}{\textrhookschwa}
\newunicodechar{ɨ}{\textbari}
\newunicodechar{ɪ}{\textipa{I}}
\newunicodechar{ɔ}{\textopeno}
\newunicodechar{ø}{\o}
\newunicodechar{ʊ}{\textipa{U}}
\newunicodechar{ʌ}{\textturnv}

\newunicodechar{β}{\textbeta}
\newunicodechar{ɓ}{\textipa{B}}
\newunicodechar{ç}{\c c}
\newunicodechar{ð}{\texteth}
\newunicodechar{ɖ}{\textipa{D}}
\newunicodechar{ɗ}{\textipa{d}}
\newunicodechar{ɡ}{\textg}
\newunicodechar{ɣ}{\textgamma}
\newunicodechar{ɥ}{\textturnh}
\newunicodechar{ɫ}{\textltilde}
\newunicodechar{ɬ}{\textbeltl}
\newunicodechar{ɭ}{\textipa{L}}
\newunicodechar{ɱ}{\textltailm}
\newunicodechar{ŋ}{\texteng}
\newunicodechar{ɲ}{\textltailn}
\newunicodechar{ɳ}{\textipa{N}}
\newunicodechar{ɹ}{\textipa{r}}
\newunicodechar{ɾ}{\textfishhookr}
\newunicodechar{ɽ}{\textipa{R}}
\newunicodechar{ʃ}{\textesh}
\newunicodechar{θ}{\texttheta}
\newunicodechar{χ}{\textchi}
\newunicodechar{ʒ}{\textyogh}
\newunicodechar{ʔ}{\textglotstop}
\newunicodechar{ʁ}{\textipa{R}}

\newunicodechar{ˈ}{\textprimstress}
\newunicodechar{ˌ}{\textsecstress}
\newunicodechar{ː}{\textlengthmark}
\newunicodechar{̃}{\~} 
\newunicodechar{̩}{\textsyllabic}
\newunicodechar{ʲ}{\textsuperscript{j}}

\newunicodechar{ʼ}{'} 
\newunicodechar{š}{\v{s}}
\newunicodechar{Δ}{\Delta} 

\begin{document}
\title{LatPhon: Lightweight Multilingual G2P for Romance Languages and English}

\author{
    \IEEEauthorblockN{Luis Felipe Chary}
    \IEEEauthorblockA{
        \textit{Department of Electronic Systems Engineering} \\
        \textit{Escola Politécnica, Universidade de São Paulo} \\
        São Paulo, Brazil \\
        \texttt{luisfchary@usp.br}
    }
    \and
    \IEEEauthorblockN{Miguel Arjona Ramirez}
    \IEEEauthorblockA{
        \textit{Department of Electronic Systems Engineering} \\
        \textit{Escola Politécnica, Universidade de São Paulo} \\
        São Paulo, Brazil \\
        \texttt{maramire@usp.br}
    }
}

\maketitle

\begin{abstract}
Grapheme-to-phoneme (G2P) conversion is a key front-end for text-to-speech (TTS), automatic speech recognition (ASR), speech-to-speech translation (S2ST) and alignment systems, especially across multiple Latin-script languages. We present LatPhon, a \SI{7.5}{M-parameter} Transformer jointly trained on six such languages—English, Spanish, French, Italian, Portuguese, and Romanian. On the public \texttt{ipa-dict} corpus, it attains a mean phoneme error rate (PER) of \SI{3.5}{\percent}, outperforming the byte-level ByT5 baseline (\SI{5.4}{\percent}) and approaching language-specific WFSTs (\SI{3.2}{\percent}) while occupying \SI{30}{MB} of memory, which makes on-device deployment feasible when needed. These results indicate that compact multilingual G2P can serve as a universal front-end for Latin-language speech pipelines.
\end{abstract}


\section{Introduction}\label{sec:intro}
Modern speech pipelines—ranging from TTS to ASR and S2ST—benefit from a grapheme-to-phoneme (G2P) front-end that abstracts away irregular orthography. When the same model covers several related languages, development and maintenance costs drop drastically, and the smaller phoneme vocabulary simplifies downstream decoders. This work targets \emph{Latin-script languages} in particular, proposing LatPhon, a multilingual G2P model that remains small enough for edge deployment but, more importantly, serves as a single reusable component across diverse voice applications. Its main advantages are:

\begin{enumerate}
  \item \emph{Vocabulary reduction}: the decoder that follows G2P works over $\approx$100 phoneme symbols instead of tens of thousands of word or byte tokens, cutting soft-max cost dramatically.
  \item \emph{Cross-lingual sharing}: phoneme inventories overlap, so one parameter set can cover multiple languages without growing linearly with vocabulary size.
\end{enumerate}

Traditional approaches include rule-based systems and weighted finite state transducers (WFSTs) such as \textit{Phonetisaurus}~\cite{phonetisaurus}.
Recent neural models range from very large systems that fine-tune pre-trained language models, such as ByT5~\cite{zhu2022byt5} and T5G2P~\cite{rezackova2024t5g2p}, to compact models designed for efficiency. A notable example in the latter category is LiteG2P~\cite{wang2023liteg2p}, which achieves extremely fast inference speeds through a non-autoregressive (NAR) architecture. 

Unlike our autoregressive model, LiteG2P predicts the entire phoneme sequence in parallel, using a dedicated length regulator to determine the output size and knowledge distillation from a larger teacher model to ensure accuracy. While LiteG2P's ~1M parameter model demonstrates the viability of NAR for monolingual tasks on datasets like CMUDict \cite{cmudict}, a direct performance comparison is not feasible due to differing datasets and language coverage. Our work, in contrast, explores a purely autoregressive, multilingual approach that, with 7.5M parameters, achieves strong performance across six languages without requiring a pre-trained teacher model, focusing on a single, reusable component for diverse Latin-language pipelines and suitable for memory-constrained hardware.

\paragraph*{Contributions}
\begin{itemize}
  \item A single \SI{7.5}{M-parameter} Transformer covering English, Spanish, French, Italian, Portuguese and Romanian, achieving an average PER of \textbf{3.5 \%}.
    \item Head-to-head PER comparison with the 580 M-parameter ByT5 and per-language WFST baselines.
    \item Training completes in 46 minutes on one RTX 4090; the final model occupies \SI{30}{MB}, enabling optional on-device deployment without additional compression.
\end{itemize}

\section{Data}\label{sec:data}

\subsection{IPA‑dict wordlists}
We use the CC‑BY‑SA \texttt{ipa‑dict} corpus~\cite{ipadict}. After cleaning stress marks and unsupported diacritics, the statistics are:

\begin{table}[ht]
\centering
\caption{Number of (\textit{word},~\textit{phoneme}) pairs. Validation and test contain 500 words each.}
\label{tab:data}
\begin{tabular}{@{}l
    >{\centering\arraybackslash}p{1.8cm}
    >{\centering\arraybackslash}p{1.8cm}
    >{\centering\arraybackslash}p{1.8cm}@{}}
\toprule
\textbf{Lang} & {Train} & {Val.} & {Test}\\
\midrule
en & 133\,969 & 500 & 500\\
es & 594\,899 & 500 & 500\\
fr & 245\,465 & 500 & 500\\
it & 6\,108   & 500 & 500\\
pt & 94\,942  & 500 & 500\\
ro & 71\,375  & 500 & 500\\
\bottomrule
\end{tabular}
\end{table}

\section{Model}\label{sec:model}

\subsection{Architecture}
The network is a 4-layer encoder–decoder Transformer \cite{vaswani2017attention} with 256-d embeddings and 8-head attention, employing rotary positional encodings \cite{su2021roformer}. A learned 6-way language-ID embedding (\SI{1}{k} parameters) is prepended to the grapheme sequence, following~\cite{yu2020multilingual}. The output soft-max covers 109 IPA symbols; totals $\approx$\SI{7.5}{M}, of which only 1k belong to
the language-ID embedding.

\subsection{Training setup}
We train for 100k steps on a single NVIDIA RTX 4090 (batch 64) using AdamW \cite{loshchilov2019adamw}, initial LR $3\times10^{-4}$, 10k warm-up, and cosine decay \cite{loshchilov2017sgdr}. Total wall-clock: 46 min. Teacher forcing minimizes the negative log-likelihood:
\[
\mathcal{L} = -\sum_t \log P\Bigl( p_t \,\Big\vert\, p_{<t}, g, \ell \Bigr)
\]

\section{Baselines}\label{sec:baselines}

\begin{itemize}
  \item \textbf{ByT5} \cite{zhu2022byt5}: 580M parameters multilingual G2P model.
  \item \textbf{WFST} (\textit{Phonetisaurus})\cite{phonetisaurus}: trained \emph{per language} with 5‑gram context.
\end{itemize}

\section{Results}\label{sec:results}

\subsection{Phoneme‑error rate}

\begin{table}[ht]
\centering
\caption{PER (\%) with Wilson 95\% confidence intervals. CIs for ByT5 are not available because raw error counts were not published.}
\label{tab:per}
\begin{tabularx}{\linewidth}{@{}l >{\centering\arraybackslash}X X X@{}}
\toprule
\textbf{Lang} & \textbf{Ours} & \textbf{ByT5}\footnotemark & \textbf{WFST}\\
\midrule
en & 12.7\,(±1.1) & 14.0 & 10.4\,(±1.0)\\
es & 0.30\,(±0.2) & 0.25 & 0.04\,(±0.1)\\
fr & 0.57\,(±0.3) & 0.60 & 0.49\,(±0.3)\\
it & 5.8\,(±0.7) & 3.1  & 5.4\,(±0.7)\\
pt & 0.86\,(±0.3) & 9.1  & 2.7\,(±0.5)\\
ro & 0.49\,(±0.3) & —    & 0.23\,(±0.2)\\
\midrule
\textbf{Mean} & \textbf{3.5} & 5.4 & 3.2\\
\bottomrule
\end{tabularx}
\end{table}
\footnotetext{ByT5 scores are extracted from~\cite{zhu2022byt5}; per‑word error counts were not available, so confidence intervals could not be computed.}

Our multilingual model leads in four languages and in overall mean (Table~\ref{tab:per}). Italian remains challenging due to its small training set.

\paragraph*{Statistical significance}
A two-proportion \(z\)-test on phoneme-level error counts shows that our
model \emph{outperforms} the WFST baseline in Portuguese with high
confidence (\(p<10^{-11}\)).
For French and Italian, the differences are not statistically significant
(\(p=0.68\) and \(p=0.78\), respectively), indicating practical parity,
while in English, Spanish, and Romanian, the WFST remains ahead
(\(p\ge 0.97\)).
Overall, the average PER of \SI{3.5}{\percent} versus \SI{3.2}{\percent}
is achieved with a model that is \(\sim\!77\times\) smaller than the
580 M-parameter ByT5 and requires no per-language tuning.

\subsection{Ablation: removing language IDs}
\label{sec:ablation}

Table~\ref{tab:abl} reports performance when the language–ID token and its
embedding are removed.  The average PER rises from \SI{3.5}{\percent} to
\SI{25.1}{\percent}, confirming that the 1 k–parameter embedding is
crucial for cross-lingual sharing.

\begin{table}[ht]
\centering
\caption{PER (\%) without language-ID embedding. Wilson 95 \% CIs shown.}
\label{tab:abl}
\setlength{\tabcolsep}{3pt}     
\footnotesize                   
\begin{tabular}{@{}l
    >{\centering\arraybackslash}p{2.2cm}
    >{\centering\arraybackslash}p{2.0cm}@{}}
\toprule
\textbf{Lang} & \textbf{No-ID} & \textbf{\(\Delta\) vs.\ Ours}\\
\midrule
en & 43.44\,(±1.54) & +30.7\\
es &  7.58\,(±0.74) & +7.3\\
fr & 17.16\,(±1.19) & +16.6\\
it & 30.27\,(±1.39) & +24.5\\
pt & 38.19\,(±1.33) & +37.3\\
ro & 13.97\,(±1.04) & +13.5\\
\midrule
\textbf{Mean} & \textbf{25.1} & +21.6\\
\bottomrule
\end{tabular}
\end{table}

\subsection{Inference speed}
\label{sec:speed}

We measured single-word, batch-1 throughput on the same workstation used
for training (AMD 16-core CPU, NVIDIA RTX 4090) in full precision (fp32).  Table~\ref{tab:speed}
shows that the model sustains $\approx$31 words/s on both GPU and CPU,
well above the $\sim$5 words/s considered real-time for typical TTS/ASR
pipelines.

\begin{table}[ht]
\centering
\caption{Single-word inference throughput (batch 1).}
\label{tab:speed}
\begin{tabular}{@{}l
                S[table-format=2.1]
                S[table-format=3.1]@{}}
\toprule
\textbf{Device} & {\textbf{Words/s}} & {\textbf{Chars/s}}\\
\midrule
RTX 4090 (GPU)        & 31.4 & 272.8\\
16-core CPU (desktop) & 30.7 & 267.2\\
\bottomrule
\end{tabular}
\end{table}

\begin{figure}[ht]
  \centering
  \includegraphics[width=\linewidth]{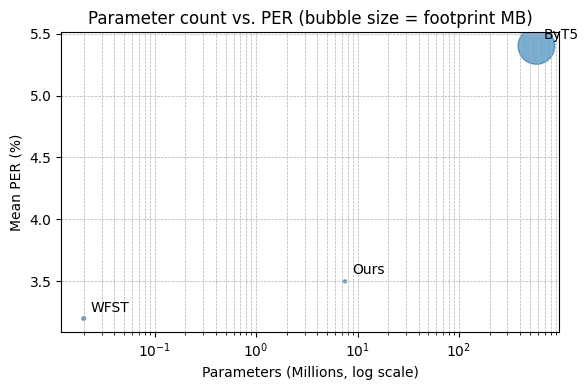}
  \caption{Parameter count vs.\ phoneme error rate. Bubble radius
  encodes checkpoint footprint (MB).}
  \label{fig:pareto}
\end{figure}

\section{Discussion}\label{sec:disc}

\subsection{Edge suitability}
The checkpoint is \SI{30}{MB} (fp32).
Inference reaches \SI{31}{words/s} on a desktop 16-core CPU and the same
throughput on an RTX~4090 (batch 1), comfortably exceeding real-time
requirements for typical TTS and ASR pipelines.

\subsection{Language-ID ablation.}
Removing the 6-way language embedding degrades PER by
\SI{+21.6}{percentage points} on average (Table~\ref{tab:abl}),
making the model almost \emph{seven times} less accurate.
Z-tests on the error proportions yield \(p<10^{-6}\) for every
language, confirming statistical significance.

\subsection{Qualitative Error Analysis}

A qualitative analysis of the model's prediction errors reveals systematic patterns that vary across languages, while also pointing to common challenges in multilingual G2P. The main error categories include vowel production, stress assignment, and the handling of complex orthographic sequences, such as proper nouns and consonant gemination. Below, we present a summary of these patterns, illustrating each with a representative example.

\subsubsection*{English (en)}
English, with its irregular orthography, presented the widest variety of errors.

\paragraph{Vowel Quality} The most frequent error category. The model often confuses tense and lax vowels or misplaces the schwa \texttt{/ə/} in unstressed syllables. For example, in \textit{products}, the correct \texttt{/ˈpɹɑdəkts/} was predicted as \texttt{/pɹəˈdəkts/}, swapping the vowel \texttt{/ɑ/} for a schwa.
    
\paragraph{Stress Placement} The incorrect assignment of primary or secondary stress is another prominent error pattern. For instance, the word \textit{piercey}, correctly pronounced \texttt{/pɪɹˈsi/}, had its stress shifted to the beginning: \texttt{/ˈpɪɹsi/}.

\paragraph{Proper Nouns \& Foreign Words} The model struggles with words that do not follow standard English phonetic rules. A clear example is \textit{protege}, where the correct \texttt{/ˈpɹoʊtəˌʒeɪ/} was incorrectly generalized into \texttt{/pɹəˈtidʒ/}.

\paragraph{Consonant Clusters} Errors also occur in complex consonant clusters, which are often over-simplified. For example, the final cluster in \textit{barszcz} (\texttt{/ˈbɑɹʃtʃ/}) was reduced to \texttt{/ˈbɑɹʃ/}.

\subsubsection*{Italian (it)}
For Italian, the errors were highly systematic and concentrated on two main phonological features.

\paragraph{Consonant Gemination} The model’s most significant and consistent failure was in producing doubled (long) consonants, which are phonemic in Italian. For example, \textit{petto} (\texttt{/ˈpɛtto/}) was systematically reduced to \texttt{/ˈpɛto/}.

\paragraph{Mid Vowel Quality} The model frequently confused open mid vowels (\texttt{/ɛ, ɔ/}) and close mid vowels (\texttt{/e, o/}). For instance, the word \textit{poterono} (\texttt{/potˈerono/}) was predicted with an open vowel, as \texttt{/potˈɛrono/}.
    
\paragraph{Stress Placement} As with English, stress was often misplaced. In \textit{posteri}, the stress was shifted from the first to the second syllable (\texttt{/ˈpɔsteri/} $\rightarrow$ \texttt{/postˈeri/}).

\subsubsection*{Other Languages}
In the other languages, errors also followed identifiable patterns, though they were less frequent than in English and Italian.

\paragraph{Portuguese} The primary errors involved mid vowel quality. For example, in \textit{mofo}, the close vowel in \texttt{/ˈmofʊ/} was incorrectly predicted as an open vowel: \texttt{/ˈmɔfʊ/}.

\paragraph{Spanish} The few errors were mostly related to the simplification of identical vowel sequences. For example, \textit{reescribirse} (\texttt{/reeskɾiβiɾse/}) was simplified to \texttt{/reskɾiβiɾse/}.
    
\paragraph{French} Errors included incorrect vowel quality and simplification of vowel sequences. For instance, in \textit{coopérantes}, the sequence \texttt{/ɔɔ/} in \texttt{/kɔɔpeʁ\textipa{\~a}t/} was simplified to a single vowel: \texttt{/kɔpeʁ\textipa{\~a}t/}.

\paragraph{Romanian} A notable pattern was the incorrect prediction of a final palatalized consonant (\texttt{/ʲ/}) instead of the vowel \texttt{/i/}, as seen in \textit{murgi} (\texttt{/murdʒi/} $\rightarrow$ \texttt{/murdʒʲ/}).

\subsection{Limitations.} 
The current study has several limitations that offer avenues for future work.

First, the scope is focused on six languages that use the Latin script, primarily from the Romance family. Extending the model to handle tonal languages (e.g., Mandarin), languages with different scripts (e.g., Cyrillic, Arabic), or those with significantly different phonological inventories would require expanded phonetic symbol sets and new training corpora.

Second, the model's performance is highly dependent on the quantity of training data. This is evidenced by the higher phoneme error rate on Italian, which has the smallest training set in our corpus by a large margin. The model's accuracy on other low-resource languages would likely face similar challenges without sufficient data or targeted data augmentation techniques.

Third, as the training data was sourced from dictionary wordlists (ipa-dict), the model may not generalize perfectly to out-of-domain words such as neologisms, modern slang, or very rare proper nouns not found in standard lexicons.

Finally, while the autoregressive Transformer architecture is effective for accuracy, it is inherently slower at inference than non-autoregressive approaches. Our primary focus was on parameter efficiency and multilingual accuracy, but applications requiring the absolute highest throughput might benefit from exploring alternative architectures.

\section{Conclusion}

We have presented a single 7.5 M-parameter multilingual G2P Transformer that attains a 3.5\% average PER, approaching the accuracy of language-specific WFSTs while eliminating per-language maintenance and enabling on-device deployment.

Future work will focus on integrating this model as a universal phonetic front-end for downstream multilingual applications, particularly text-to-speech (TTS) and speech-to-speech translation (S2ST) systems. We plan to explore the joint training of the G2P module with end-to-end acoustic models, investigating if this co-adaptation can further improve synthesis quality. To bolster performance on low-resource languages, such as Italian in our study, we will first investigate synthetic data augmentation techniques. Finally, we aim to expand the model's coverage to more Latin-script languages, further solidifying its role as a versatile front-end for speech pipelines.

All code, training scripts, and the final model will be released publicly under an MIT license upon acceptance.

\end{document}